\pdfoutput=1

\documentclass[11pt]{article}

\usepackage{acl}

\usepackage{times}
\usepackage{latexsym}
\usepackage[export]{adjustbox}
\usepackage[T1]{fontenc}

\usepackage[utf8]{inputenc}

\usepackage{microtype}

\usepackage{inconsolata}

\usepackage{graphicx} 
\usepackage{enumitem}
\usepackage{todonotes}
\usepackage{xcolor,colortbl}
\usepackage{multirow}
\usepackage{booktabs}
\usepackage{caption}
\usepackage{subcaption}

\newcommand\nnfootnote[1]{%
  \begin{NoHyper}
  \renewcommand\thefootnote{}\footnote{#1}%
  \addtocounter{footnote}{-1}%
  \end{NoHyper}
}

\definecolor{brilliantlavender}{cmyk}{0, 0.2235, 0, 0.1}
\definecolor{celeste}{cmyk}{0.3922, 0.0353, 0, 0.1}

\newcommand{\narrowtextsc}[1]{\textls[-50]{\textsc{#1}}}
\newcommand{\lm}[1]{\texttt{#1}}
\newcommand{\sys}[1]{\narrowtextsc{#1}}
\newcommand{\data}[1]{\textsf{#1}}
\newcommand{\affilsup}[1]{\rlap{\textsuperscript{\normalfont#1}}}

\title{LLMCheckup: \\Conversational Examination of Large Language Models \\via Interpretability Tools and Self-Explanations}

\author{Qianli Wang\affilsup{1,2}
    \qquad
    Tatiana Anikina$^*$\affilsup{1,3}
    \qquad
    Nils Feldhus$^*$\affilsup{1}
    \\
    \textbf{Josef van Genabith}\affilsup{1,3}
    \qquad
    \textbf{Leonhard Hennig}\affilsup{1}
    \qquad 
    \textbf{Sebastian Möller}\affilsup{1,2}
    \\
    $^1$German Research Center for Artificial Intelligence (DFKI) \\
    $^2$Technische Universit\"at Berlin, Germany \\
    $^3$Saarland Informatics Campus, Saarbrücken, Germany \\
    \texttt{\{firstname.lastname\}@dfki.de} \\
}

\begin{document}
\maketitle

\nnfootnote{*Equal contribution}

\begin{abstract}
Interpretability tools that offer explanations in the form of a dialogue have demonstrated their efficacy in enhancing users' understanding \cite{slack-2023-talktomodel, shen-2023-convxai}, as one-off explanations may fall short in providing sufficient information to the user.
Current solutions for dialogue-based explanations, however, often require external tools and modules and are not easily transferable to tasks they were not designed for.
With \sys{LLMCheckup}\footnote{
\url{https://github.com/DFKI-NLP/LLMCheckup}}, we present an easily accessible tool that allows users to chat with any state-of-the-art large language model (LLM) about its behavior. 
We enable LLMs to generate explanations and perform user intent recognition without fine-tuning, by connecting them with a broad spectrum of Explainable AI (XAI) methods, including white-box explainability tools such as feature attributions, and self-explanations (e.g., for rationale generation).
LLM-based (self-)explanations are presented as an interactive dialogue that supports follow-up questions and generates suggestions. 
\sys{LLMCheckup} provides tutorials for operations available in the system, catering to individuals with varying levels of expertise in XAI and supporting multiple input modalities. 
We introduce a new parsing strategy 
that substantially enhances the user intent recognition accuracy of the LLM. 
Finally, we showcase \sys{LLMCheckup} for the tasks of fact checking and commonsense question answering.
\end{abstract}

\begin{figure}[t!]
    \centering
    \includegraphics[width=\linewidth]{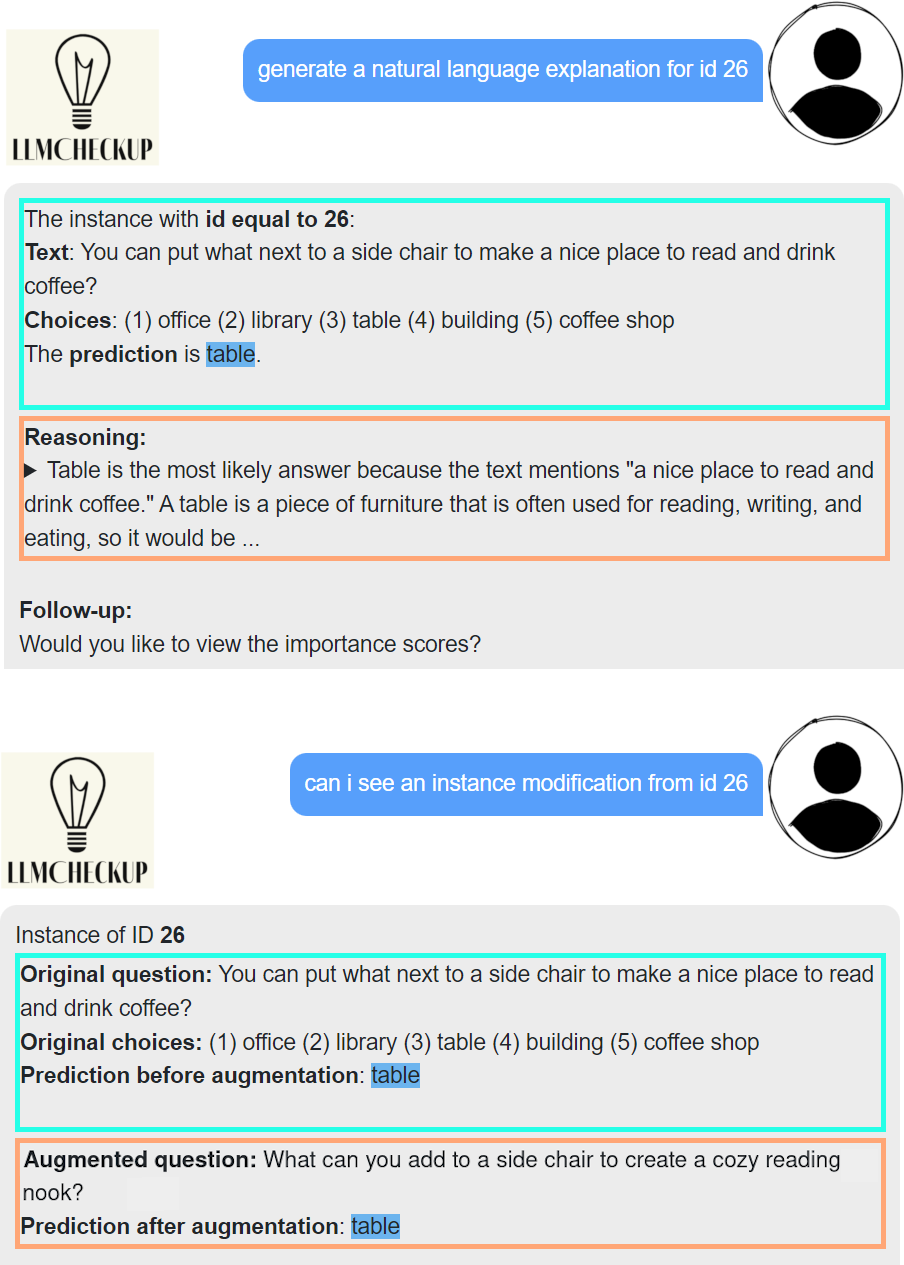}
    \caption{\sys{LLMCheckup} dialogue with data augmentation and rationalization operations on a commonsense question answering task (\data{ECQA}). 
     Boxes (not part of the actual UI) indicate the original instance from the dataset as well as its prediction (cyan) and the explanation requested by the user (orange).
    }
    \label{fig:example}
\end{figure}

\section{Introduction}
To unravel the black box nature of deep learning models for natural language processing, a diverse range of explainability methods have been developed \cite{ribeiro-2016-lime, madsen-2022-post-hoc, wiegreffe-etal-2022-reframing}. Nevertheless, practitioners often face difficulties in effectively utilizing explainability methods, as they may not be aware of which techniques are available or how to interpret results provided. There has been a consensus within the research community that moving beyond one-off explanations and embracing conversations to provide explanations is more effective for model understanding \cite{lakkaraju-2022-rethinking, feldhus-etal-2023-interrolang, zhang-2023-may-i-ask} and helps mitigate the limitations associated with the effective usage of explainability methods to some extent \cite{ferreira-2020-what-are-people-doing, slack-2023-talktomodel}. 

\begin{figure*}[t!]
    \centering
    \resizebox{\textwidth}{!}{%
        \includegraphics{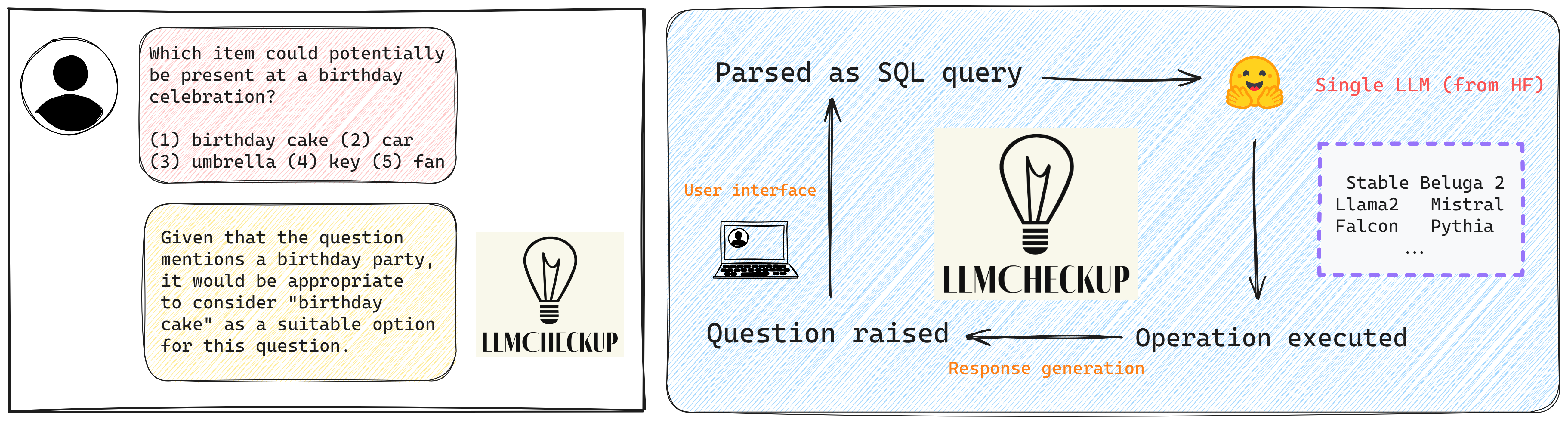}
    }
    \caption{
    On the left, a dialogue example asking for explanation in natural language about a \data{ECQA}-like customized question. The workflow of \sys{LLMCheckup} is shown on the right side.
    }
    \label{fig:architecture}
\end{figure*}

In the field of NLP, two dialogue-based interpretability tools, \sys{InterroLang} \cite{feldhus-etal-2023-interrolang} and \sys{ConvXAI} \cite{shen-2023-convxai}, have been introduced. Both tools employ multiple, separately fine-tuned LMs to parse user intents and dedicated external LMs to provide explanations.

By contrast, our framework, \sys{LLMCheckup}, only requires a single LLM and puts it on ``quadruple duty'': (1) Analyzing users' (explanation) requests (\S \ref{subsec:architecture}, \S \ref{subsec:parsing}), (2) performing downstream tasks (\S \ref{sec:use_case}), (3) providing explanations for its outputs (\S \ref{sec:exp_methods}), and (4) responding to the users in natural language (\S \ref{subsec:interface}).
Instead of using many different LMs to explain the behavior of another LLM,
\sys{LLMCheckup} allows us to directly employ the same LLM used for user intent recognition to self-explain its own behavior. The advantage of a single-model approach is that it simplifies the engineering aspect of building an XAI system without multiple external modules and separately fine-tuned models. 
At the same time, we consistently achieve good performance even with a single model, as modern LLMs are  very powerful and can handle a wide range of tasks including user intent recognition and explanation generation.
Thus, \sys{LLMCheckup} provides a unified and compact framework that is useful for future user studies in the context of human-computer interaction and explainability.
%

\section{\textsc{LLMCheckup}}

\sys{LLMCheckup} is an interface for chatting with any LLM about its behavior. We connect several white-box and black-box interpretability tools (\S \ref{sec:exp_methods}), s.t. \sys{LLMCheckup} takes into account model internals, datasets and documentation for generating self-explanations. User requests for explanations are recognized via a text-to-SQL task performed by the LLM under investigation (\S \ref{subsec:architecture}-\ref{p:parsing}).

We showcase a short dialogue between the user and \sys{LLMCheckup} in Figure \ref{fig:example} and a longer dialogue featuring different operations in  Appendix \ref{sec:examples}. \sys{LLMCheckup} can answer various questions related to the data as well as the model's behavior. For example, in Figure \ref{fig:example} the user is interested in the rationale for a specific prediction and the model generates an explanation to justify the assigned label. \sys{LLMCheckup} also suggests to have a look at another related operation (token-level importance scores) that can help explain model's behavior (\S \ref{p:folow-up}), but the user asks for a modified (augmented) version of the same instance instead. As a result, the model paraphrases the original question which can be then treated as a new sample and the user can further examine it by using the custom input functionality of \sys{LLMCheckup} (\S \ref{p:cutom_inputs}).

\subsection{System architecture}
\label{subsec:architecture}
Figure~\ref{fig:architecture} illustrates the interaction flow of \sys{LLMCheckup}. When a user asks a question, it will be parsed as an SQL-like query by the LLM. E.g., the first user question in Figure~\ref{fig:example} will be parsed as \texttt{filter id 26 and rationalize}. The corresponding parsed operation (i.e., \texttt{filter} and \texttt{rationalize} in our example, see Table~\ref{tab:ops} for the full list of operations) will then be matched and executed. 
For response generation, the explanation provided by the underlying interpretability method is converted into a natural language output using a template-based approach \cite{slack-2023-talktomodel,feldhus-etal-2023-interrolang} and is then displayed to the user.

\begin{table*}[ht!]
    \centering
    \renewcommand*{\arraystretch}{1.05}
    \footnotesize
    \scalebox{0.75}{
        \begin{tabular}{|lp{3.5cm}|p{5.5cm}|}
        \toprule
        \multirow{2}{*}{\centering \rotatebox[origin=c]{90}{\textbf{Filter}}}
        & \texttt{\textcolor{blue}{filter}(id)}
        & Access single instance by its ID \\
        
        & \texttt{\textcolor{blue}{includes}(token)}
        & Filter instances by token occurrence \\
        & & \\

        \midrule 
        \multirow{4}{*}{\centering \rotatebox[origin=c]{90}{\textbf{Prediction}}}
        & \texttt{\textcolor{blue}{predict}(instance)$^*$} 
        & Get the prediction for the given instance \\
        
        & \texttt{\textcolor{blue}{randompredict}(number)} 
        & Precompute a subset of instances at random\\
        
        & \texttt{\textcolor{blue}{mistakes show|count}(subset)} 
        & Count or show incorrectly predicted instances\\
        
        & \texttt{\textcolor{blue}{score}(subset, metric)} 
        & Determine the relation between predictions and labels\\

        \midrule 
        \multirow{3}{*}{\centering \rotatebox[origin=c]{90}{\textbf{Data}}}
        & \texttt{\textcolor{blue}{show}(list)} 
        & Showcase a list of instances\\
        
        & \texttt{\textcolor{blue}{countdata}(list)} 
        & Count number of instances in the dataset\\
        
        & \texttt{\textcolor{blue}{label}(dataset)} 
        & Describe the label distribution across the dataset\\

        \midrule 
        \multirow{2}{*}{\centering \rotatebox[origin=c]{90}{\textbf{Meta}}}
        & \texttt{\textcolor{blue}{data}()} 
        & Information related to the dataset\\
        
        & \texttt{\textcolor{blue}{model}()} 
        & Metadata of the model\\
        \bottomrule
        \end{tabular}
    }
    \scalebox{0.75}{
    \begin{tabular}{|l p{3.3cm}|p{5.4cm}|}
        \toprule
        \multirow{3}{*}{\centering \rotatebox[origin=c]{90}{\textbf{About}}}
        & \texttt{\textcolor{blue}{function}()} 
        & Inform about the functionality of the system\\
        
        & \texttt{\textcolor{blue}{self}()} 
        & Self-introduction of \sys{LLMCheckup}\\
        & \texttt{\textcolor{blue}{qatutorial}(op\_name)} 
        & Provide explanation for the supported operations (tutorial)  \\

        \midrule 
        \multirow{2}{*}{\centering \rotatebox[origin=c]{90}{\textbf{Explain}}} 
        & \texttt{\textcolor{blue}{nlpattribute}(inst., topk, method\_name)$^*$} 
        & Provide feature attribution scores  \\
        & \texttt{\textcolor{blue}{rationalize}(instance)$^*$} 
        & Explain the output in natural language  \\

        \midrule 
        \multirow{2}{*}{\centering \rotatebox[origin=c]{90}{\textbf{NLU}}} 
        & \texttt{\textcolor{blue}{keywords}()} 
        & Show common keywords in the data  \\
        & \texttt{\textcolor{blue}{similarity}(instance, number)$^*$} 
        & Output top $k$ similar instances in the dataset  \\

        \midrule 
        \multirow{2}{*}{\centering \rotatebox[origin=c]{90}{\textbf{Pert.}}} 
        & \texttt{\textcolor{blue}{cfe}(instance)$^*$} 
        & Generate counterfactuals  \\
        & \texttt{\textcolor{blue}{augment}(instance)$^*$} 
        & Augment the input text  \\

        \midrule 
        \multirow{2}{*}{\centering \rotatebox[origin=c]{90}{\textbf{Logic}}}
        & \texttt{\textcolor{blue}{and}(op1, op2)} 
        & Concatenation of multiple operations \\ 
        & \texttt{\textcolor{blue}{or}(op1, op2)} 
        & Selection of multiple filters \\
        \bottomrule
        \end{tabular}
        }
    \caption{
    All operations (mappings between a partial SQL-type query and a function) facilitated by \sys{LLMCheckup}, including all explainability methods mentioned in \S \ref{sec:exp_methods} and other supplementary operations.
    Operations marked with ($^*$) support the use of custom inputs (see more details in App.~\ref{app:ops}).
    }
    \label{tab:ops}
\end{table*}

\subsection{Parsing}
\label{p:parsing}
To recognize users' intents, the deployed LLM transforms a user utterance into a SQL-like query.
The SQL-based approach is needed to formally represent the available operations (see Table \ref{tab:ops}) and their ``semantics'' including all necessary attributes.
For user intent recognition, we employ two methods: Guided Decoding and Multi-prompt Parsing. 

\subsubsection{Guided Decoding}
Guided Decoding (GD) ensures that the generated output adheres to predefined grammatical rules and constraints \cite{shin-etal-2021-constrained} and that parses of the user requests align with predefined operation sets \cite{slack-2023-talktomodel}. GD is generally more suitable for smaller LMs, since in-context learning may encounter instability attributed to the fluctuations in the order of provided demonstrations, and the formats of prompts \cite{ma-2023-fairness-guided}. 

\subsubsection{Multi-prompt Parsing}
As an alternative to GD, we propose and implement a novel Multi-prompt Parsing (MP) approach. While GD pre-selects prompts based on the embedding similarity with user input, the model does not see all the available operations at once and the pre-selection may not include the examples for the actual operation required. 
With MP, we test whether showing all possible operations in a simplified format 
(i.e., without any attributes such as instance ID or number of samples) and then additionally prompting the model to fill in more fine-grained attributes can improve performance.

As a first step, MP queries the model about the main operation (see list of operations in Table~\ref{tab:ops}). Next, depending on the chosen operation, MP selects the operation-specific prompts with 2-7 demonstrations\footnote{The number of demonstrations depends on the difficulty of operation, e.g., how many attributes it may have.} (user query and parsed outputs examples) to generate the full parses that may include several attributes. E.g., for the user input \textit{"What are the feature attributions for ID 42 based on the integrated gradients?"}, we start by generating \texttt{nlpattribute} and then augment the parse with the second prompt and transform it into \texttt{filter id 42 and nlpattribute integrated\_gradient}.

Since the output of the model is not constrained, unlike in GD, in the MP setting we need to check whether the model's output matches any of the available operations and if there is no exact match we employ \lm{SBERT}\footnote{\url{https://huggingface.co/sentence-transformers/all-mpnet-base-v2}} to find the best match based on the embedding similarity. 
We also implement checks to avoid possible hallucinations, e.g., if the model outputs an ID that is not present in the input we remove it from the parser output. \S\ref{subsec:parsing} evaluates the performance of both parsing approaches. 


\begin{figure*}[ht!]
\centering
\resizebox{\textwidth}{!}{
\begin{minipage}{\textwidth}
\includegraphics[width=\linewidth]{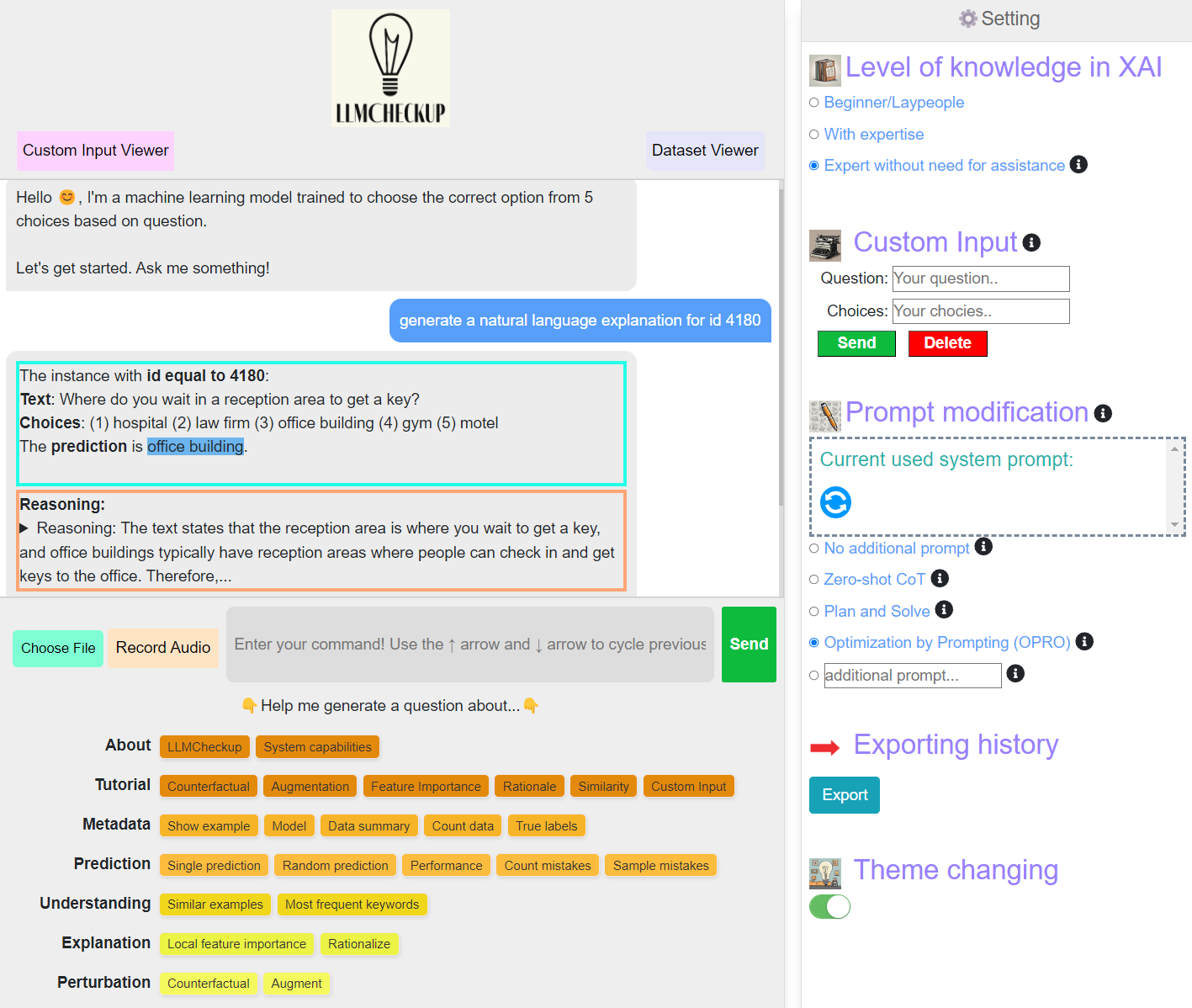}
\end{minipage}
}
\caption{\sys{LLMCheckup} interface with welcome message, 
free-text rationale and sample generator buttons. Expert XAI level and \textit{OPRO} strategy are selected. For example multi-turn dialogues, see Table~\ref{tab:examples_1} and Table~\ref{tab:examples_2}.}
\label{fig:interface}
\end{figure*}

\subsection{Interface}
\label{subsec:interface}
\sys{LLMCheckup} provides a user interface (Figure~\ref{fig:interface}) including a chat window to enter questions and settings on the right panel, including XAI expertise level selection, custom inputs, prompt editor and export functionality for the chat history. It is implemented in Flask and can be run as a Docker container.
\sys{LLMCheckup} provides a chat window \cite{slack-2023-talktomodel}, a dataset viewer \cite{feldhus-etal-2023-interrolang}, a custom input history viewer and question suggestions for different operations. Together, these UI elements facilitate dataset exploration and provide sample questions for all available operations to inspire users to come up with their own questions. 


On the right side of the window, there is a Prompt Editor with different options for prompt modification (\S \ref{sec:blackbox}). The icons associated with each strategy describe them in detail, including the corresponding prompts that can be appended after the default system prompt.

\subsection{Key features}
\label{subsec:key_feature}

\paragraph{Supported NLP models}
\label{p:nlp_models}
Out of the box, we include five auto-regressive LLMs representative of the current state-of-the-art in open-source NLP (as indicated in the left column of Table~\ref{tab:parsing}) available through Hugging Face \sys{Transformers} \cite{wolf-etal-2020-transformers}. 
The diverse choice of models demonstrates that our framework is generalizable and supports various Transformer-type models. 
While \lm{Falcon-1B} \cite{penedo-2023-refinedweb} and \lm{Pythia-2.8B} \cite{biderman-2023-pythia} are available for users with limited hardware resources (RAM/GPU), it is generally not recommended to use them due to their small model size, which may negatively affect performance and user perception. 
\lm{Llama2-7B} \cite{touvron-2023-llama-2} and \lm{Mistral-7B} \cite{jiang-2023-mistral} are both mid-sized with 7B parameters, while \lm{Stable Beluga 2} \cite{mukherjee-2023-orca} is a fine-tuned version of \lm{Llama2-70B}.
To facilitate the deployment of large models in a local environment, \sys{LLMCheckup} offers support for various forms of LLMs. This includes \textbf{quantized models} through \sys{GPTQ} \cite{frantar-2023-gptq}, loading models in 4-bits with the assistance of \sys{bitsandbytes} \cite{dettmers-2022-optimizers}, and the implementation of a \textbf{peer-to-peer} solution using \sys{Petals} \cite{borzunov-etal-2023-petals}, enabling efficient deployment on a custom-level GPU.

\paragraph{Tutorial}
To help non-experts get background knowledge in XAI, we introduce a tutorial functionality. It is based on prompting with different roles corresponding to levels of expertise in XAI (Figure~\ref{fig:interface}) and enables us to provide tailored meta-explanations of supported operations to individuals. 
For example, at the beginner level, we add a system prompt hinting at the expertise: \textit{``As a NLP beginner, could you explain what data augmentation is?''} (Figure~\ref{fig:subfigures}). In such a way, all users can receive meta-explanations according to their expertise.

\paragraph{Customized inputs \& prompts}
\label{p:cutom_inputs}
In comparison to \sys{TalkToModel} \cite{slack-2023-talktomodel}, which was limited to three datasets, \sys{LLMCheckup} offers users the freedom to enter custom inputs (e.g. modified original samples or even completely new data points, see the Custom Input box on the right panel in Figure \ref{fig:interface}), going beyond just querying instances from specific provided datasets. 
In addition, inspired by \sys{PromptSource} \cite{bach-etal-2022-promptsource}, a Prompt Editor (see Prompt modification section on the right panel in Figure \ref{fig:interface}) supports inserting both pre-defined and fully customized prompts, allowing the users to control how downstream tasks and rationalization (\S \ref{p:rationale}) are performed.
All custom inputs are saved and can be inspected and reused later via a dedicated custom input history viewer.

\paragraph{Suggestion of follow-up questions}
\label{p:folow-up}
To guide the user through the conversation, we implemented a suggestions mode. 
The user receives suggestions for related operations that \sys{LLMCheckup} can perform based on the dialogue context, e.g., if the user asks about the top $k$ attributed tokens for a specific sample, they will receive a suggestion to have a look at the generated rationales since both operations belong to the ``Explanation" category also displayed in the user interface. The suggestions are grouped into several categories as specified in Table~\ref{tab:ops} (see Appendix~\ref{app:follow-up} for more detail).



\subsection{Add-on features}
\paragraph{External information retrieval}
Since LLMs may sometimes generate incorrect responses \cite{welleck-2020-unlikelihood-training}, \sys{LLMCheckup} allows users to access information by conducting search through external knowledge bases, promoted by the integration of \sys{Google Search}\footnote{\url{https://github.com/Nv7-GitHub/googlesearch}} (Figure~\ref{fig:ir_covid}). 
In particular, it provides an external link that contains information relevant to the input sample(s).
Users can cross-reference the retrieved information with the provided explanations, thereby achieving a more comprehensive understanding. 

\paragraph{Multi-modal input format}
Motivated by \citet{malandri-2023-convxai}, \sys{LLMCheckup} not only accepts text input from users but also provides support for other modalities such as images and audio. 
To facilitate this, we integrate packages and models tailored to each modality. For optical character recognition (OCR), we use \sys{EasyOCR}\footnote{\url{https://github.com/JaidedAI/EasyOCR}}. For audio recognition, we employ a lightweight \lm{fairseq S2T}\footnote{\url{https://huggingface.co/facebook/s2t-small-librispeech-asr}} model \cite{wang-etal-2020-fairseq} trained on Automatic Speech Recognition (ASR).

\paragraph{Dialogue sharing}
\sys{LLMCheckup} offers the functionality to export the dialogue history between the user and the deployed LLM as a JSON file that contains the user's questions and the corresponding generated responses. This simplifies data collection and sharing of conversation logs between users.

\section{NLP explainability tools}
\label{sec:exp_methods}
While we introduce each explainability method individually,  these methods can be interconnected through follow-up questions from users or suggestions provided by \sys{LLMCheckup} to preserve context. Table~\ref{tab:examples_1} and Table~\ref{tab:examples_2} show examples of explanations for each supported explainability method by \sys{LLMCheckup}.

\subsection{White-box}

\paragraph{Feature attribution} Feature attribution methods quantify the contribution of each input token towards the final outcome. In \sys{LLMCheckup}, we deploy various auto-regressive models (\S \ref{p:nlp_models}), for which \sys{Inseq} \cite{sarti-etal-2023-inseq} is used to determine attribution scores. We support representative methods from \sys{Inseq}, including \textit{Input x Gradient} \cite{simonyan-2014-deep-inside}, \textit{Attention} \cite{bahdanau-2015-neural}, \textit{LIME} \cite{ribeiro-2016-lime}, and \textit{Integrated Gradients} \cite{sundararajan-2017-axiomatic}\footnote{Details on the \sys{Inseq} integration are described in App.~\ref{app:fa}.}.

\paragraph{Embedding analysis}
By calculating the cosine similarity between the sentence embeddings of the instances in datasets, we can 
retrieve relevant examples \cite{cer-etal-2017-semeval, reimers-gurevych-2019-sentence} and present them for contextualizing the model behavior on the input in question.

\subsection{Black-box}
\label{sec:blackbox}
\paragraph{Data augmentation}
Augmentation involves synthesizing new instances by replacing text spans of the input while preserving the semantic meaning and predicted outcomes \cite{ross-etal-2022-tailor}. Data augmentation can be achieved by LLM prompting with or without providing a few demonstrations \cite{dai-2023-auggpt}.  Alternatively, \sys{NLPAug}\footnote{\url{https://github.com/makcedward/nlpaug}} can be used to substitute input words with synonyms from \sys{WordNet} \cite{miller-1995-wordnet}. Augmented texts can offer valuable insights into model behavior on perturbation tasks and prediction differences between them and their original texts. 

\paragraph{Counterfactual generation}
Unlike data augmentation, counterfactuals manifest as input edits causing the predicted outcome to be different \cite{wu-etal-2021-polyjuice, chen-etal-2023-disco}. Counterfactuals are generated by prompting LLMs with manually crafted demonstrations.

\paragraph{Rationalization} \label{p:rationale}
Rationalization aims to provide free-text explanations that elucidate the prediction made by the model \cite{camburu-2018-e-snli,wiegreffe-etal-2022-reframing} (an example is shown in Figure~\ref{fig:example}).
The use of \textit{Chain-of-Thought} (CoT) prompting enhances the reasoning capabilities of LLMs by encouraging the generation of intermediate reasoning steps that lead to a final answer \cite{wei-2022-chain-of-thought, wang-2023-self-consistency}. Different CoT strategies can be applied depending on users' preferences, including \textit{Zero-CoT} \cite{kojima-2022-zero-shot-reasoners}, \textit{Plan-and-Solve} \cite{wang-etal-2023-plan}, and \textit{Optimization by PROmpting} (OPRO) \cite{yang-2023-llms-as-optimizers} (Figure~\ref{fig:interface}).

\section{Use cases}
\label{sec:use_case}
In this paper, we demonstrate the workflow of \sys{LLMCheckup} on two typical NLP tasks: Fact checking and commonsense question answering. Figure \ref{fig:example} and Appendix \ref{sec:examples} show sample dialogues where user asks questions regarding rationalization, data augmentation and other operations based on the \data{ECQA} data \cite{aggarwal-etal-2021-explanations} for commonsense question answering. The \sys{LLMCheckup} repository includes all the necessary configuration files for different LMs and our use cases. They can be easily adopted to many other downstream tasks, data and Transformer-type models, demonstrated in a tutorial which will be available with the camera-ready version of our repository.

\subsection{Fact checking}
The importance of fact checking has grown significantly due to the rapid dissemination of both accurate information and misinformation within the modern media ecosystem \cite{guo-etal-2022-survey}. \data{COVID-Fact} \cite{saakyan-etal-2021-covid} is a fact-checking dataset that encompasses various claims, supporting evidence for those claims, and contradictory claims that have been debunked by the presented evidence.

\subsection{Commonsense question answering}
Unlike question answering, commonsense question answering (CQA) involves the utilization of background knowledge that may not be explicitly provided in the given context \cite{ostermann-etal-2018-semeval}. The challenge lies in effectively integrating a system's comprehension of commonsense knowledge and leveraging it to provide accurate responses to questions. \data{ECQA} \cite{aggarwal-etal-2021-explanations} is a dataset designed for CQA. Each instance in \data{ECQA} consists of a question, multiple answer choices, and a range of explanations. Positive explanations aim to provide support for the correct choice, while negative ones serve to refute incorrect choices. Additionally, free-text explanations are included as general natural language justifications.


\section{Evaluation}
We conducted evaluations for parsing and data augmentation with LLMs using automated evaluation metrics\footnote{Note that our evaluation does not involve any user study, as that aspect is considered as future work and falls outside the scope of our initial focus on engineering.}. Among all the supported 
methods presented in Table~\ref{tab:ops}, we chose data augmentation as a representative operation to evaluate the performance of different LLMs.

\subsection{Parsing}
\label{subsec:parsing}
\begin{table}[t!]
    \centering
    \resizebox{\columnwidth}{!}{%
        \begin{tabular}{rrr|c}

        \toprule
        \textbf{Model} & \textbf{Size} & \textbf{Strategy} & \textbf{Accuracy} \\

        \midrule
        Nearest Neighbor & - & - & 42.24 \\
        
        \midrule
        
        \lm{Falcon} & 1B & \colorbox{celeste}{GD} & 47.41\\
        \lm{Pythia} & 2.8B & \colorbox{celeste}{GD} & 51.72\\
        \lm{Llama2} & 7B & \colorbox{celeste}{GD} & 64.71\\
        \lm{Mistral} & 7B & \colorbox{celeste}{GD} & 55.88\\
        \lm{Stable Beluga 2} & 70B & \colorbox{celeste}{GD} & 67.23 \\
        
        \midrule

        \lm{Falcon} & 1B & \colorbox{brilliantlavender}{MP} & 
        64.15\\
        \lm{Pythia} & 2.8B & \colorbox{brilliantlavender}{MP} & 75.91\\
        \lm{Llama2} & 7B & \colorbox{brilliantlavender}{MP} & 82.35\\
        \lm{Mistral} & 7B & \colorbox{brilliantlavender}{MP} & 84.87\\
        \lm{Stable Beluga 2} & 70B & \colorbox{brilliantlavender}{MP} & 88.24\\
        
        \bottomrule
        \end{tabular}
    }
    \caption{
    Exact match parsing accuracy (in \%) for different models. \colorbox{celeste}{GD} = Guided Decoding prompted by 20-shots; \colorbox{brilliantlavender}{MP} = Multi-Prompt parsing. 
    }
    \label{tab:parsing}
\end{table}
To assess the ability of interpreting user intents by LLMs, we quantify the performance of each deployed model by calculating the exact match parsing accuracy \cite{talmor-etal-2017-evaluating, yu-etal-2018-spider} on a manually created test set, which consists of a total of 119 pairs of user questions and corresponding SQL-like queries. As an additional baseline, we employ the nearest neighbor approach that relies on comparing semantic similarity.

We assess parsing accuracy of our two approaches, GD and MP (\S \ref{p:parsing}). Table~\ref{tab:parsing} shows that, as model size increases, the parsing accuracy tends to increase and MP demonstrates a notable improvement over GD. 
Despite \lm{Stable Beluga 2} having a larger size compared to 7B models, its parsing performance only marginally surpasses that of \lm{Mistral} and \lm{Llama2}. This can be partially attributed to the difficulty of the parsing task\footnote{We have a total of 21 \sys{LLMCheckup} operations displayed in Table~\ref{tab:ops} (excluding the logic operations), and many of these offer multiple options. For instance, \textit{score} operation supports \textit{$F_1$}, \textit{precision}, \textit{recall} and \textit{accuracy} matrices.} and the number of demonstrations, as larger models may require a greater number of demonstrations to fully comprehend the context \cite{li-2023-evalm}.

\begin{table}[t!]
    \centering
    \renewcommand*{\arraystretch}{0.85}
    \resizebox{\columnwidth}{!}{%
        \begin{tabular}{rc|c}

        \toprule
        \textbf{Model} & \textbf{\#max\_new\_tokens} & \textbf{Accuracy} \\
        \midrule

        \lm{Falcon} & 10 & 64.15\\
        \lm{Falcon} & 20 & 64.15\\
        \midrule
        
        \lm{Pythia} & 10 & \textbf{75.91}\\
        \lm{Pythia} & 20 & 63.03\\
        \midrule
        
        \lm{Llama2} & 10 & 74.79\\
        \lm{Llama2} & 20 & 82.35\\

        \lm{Llama2-GPTQ} & 10 & 82.63\\
        \lm{Llama2-GPTQ} & 20 & \textbf{87.40}\\
        \midrule
        
        \lm{Mistral} & 10 & \textbf{84.87}\\
        \lm{Mistral} & 20 & 71.43\\

        \lm{Mistral-GPTQ} & 10 & 78.71\\
        \lm{Mistral-GPTQ} & 20 & 68.91\\
        \midrule
        
        \lm{Stable Beluga 2} & 10 & \textbf{88.24}\\
        \lm{Stable Beluga 2} & 20 & 86.55\\
        
        \bottomrule
        \end{tabular}
    }
    \caption{
        Parsing accuracy (in \%) using \colorbox{brilliantlavender}{MP} with different number of maximum new tokens. Note that for the \lm{Llama2-7b} and \lm{Mistral-7b} models, we offer various options for quantization. In this case, we have chosen \sys{GPTQ} as the representative method.
    }
    \label{tab:max_token}
\end{table} 

Table~\ref{tab:max_token} summarizes our parsing evaluation results for different models with different number of \textit{'max\_new\_tokens`} for generation. 
\lm{Llama}-based models showed better performance with more tokens to generate compared to the rest of the models.
After looking at some generated outputs we realized that \lm{Falcon-1B} and \lm{Pythia-2.8B} are not good at extracting ids and often can only recognize the main \sys{LLMCheckup} operation. Hence, for these two models we have an additional step that extracts a potential ID from the user input and adds it to the parsed operation. As expected, larger models tend to perform better than the ones with fewer parameters. However, we also found that the quantized \lm{Llama} model outperforms its full (non-quantized) version on the parsing task. 

\subsection{Data augmentation}
\begin{table*}[hbt!]
    \centering

    \resizebox{\textwidth}{!}{%
        \begin{tabular}{rr|ccc|cc}

        \toprule
        \multicolumn{2}{r|}{\textbf{Dataset}} & \multicolumn{3}{c|}{\data{COVID-Fact}} & \multicolumn{2}{c}{\data{ECQA}}  \\
        \textbf{Model} & \textbf{Size} & \textit{Consistency} & \textit{Claim Fluency} & \textit{Evidence Fluency} & \textit{Consistency} & \textit{Question Fluency}  \\

        \midrule
        \lm{Mistral} & 7B 
            & 0.66 & 0.88 
            & 0.96 & 0.50
            & 0.76   \\ 
        \lm{Llama2} & 7B 
            & 0.65 & 0.88
            & 0.94 & 0.50
            & 0.76  \\ 
        \lm{Stable Beluga 2} & 70B 
            & 1.00 & 0.85
            & 0.96 & 1.00
            &  0.73 \\
        
        \bottomrule
        \end{tabular}
    }
    \caption{
    Consistency and fluency scores of data augmentation from three models. 
    \lm{falcon} and \lm{pythia} are not considered due to poor performance because of small model size.
    }
    \label{tab:data_augmentation}
\end{table*} 

We assess the quality of the generated augmented output based on two key aspects: (1) \textbf{consistency:} the metric represents the proportion of instances where the augmentation process does not lead to a change in the label before and after the augmentation \cite{li-2023-dail, dai-2023-auggpt}; (2) \textbf{fluency:} assesses how well the augmented output aligns with the original data in terms of semantic similarity \cite{ross-etal-2021-explaining} measured by \lm{SBERT}. Table~\ref{tab:data_augmentation} indicates that \lm{Mistral} and \lm{Llama2} exhibit comparable performance, while \lm{Stable Beluga 2} displays substantially higher consistency scores on two tasks, although it may exhibit lower fluency in certain cases. The overall performance on \data{ECQA} is relatively low compared to \data{COVID-Fact}. This difference in performance can be attributed to the increased complexity of the \data{ECQA} task. Our primary focus is to compare the performance of different LMs (Table~\ref{tab:data_augmentation}), rather than aiming for state-of-the-art results on both downstream tasks or demonstrating perfect fluency and consistency\footnote{Creating gold data is out of scope for this work, because it involves costly human annotations. For the lack of gold data, we have intentionally omitted providing a baseline.}.

\section{Discussion}
In contrast to previous dialogue-based XAI frameworks \sys{ConvXAI} \cite{shen-2023-convxai} and \sys{InterroLang} \cite{feldhus-etal-2023-interrolang}, which require a fine-tuned model for each specific use case, LLMs used in \sys{LLMCheckup} possess remarkable zero-/few-shot capabilities \cite{brown-2020-gpt-3} for effectively handling many tasks without requiring fine-tuning. Although the quality of an explanation could be enhanced with further fine-tuning, \sys{LLMCheckup} uses model outputs out of the box.

Our empirical results underline the feasibility of conversational interpretability and the usefulness of \sys{LLMCheckup} for future studies, especially human evaluation. 
We focus on the ground work in terms of engineering, implementation and user interface, for connecting the human with the model.
This provides user studies \cite{wang-2019-theory-driven-xai,feldhus-etal-2023-interrolang,zhang-2023-may-i-ask} in the future with a head start, s.t. they can spend more time on conducting their study.
We see evaluation measures for differences between users' mental models and model behavior and objective metrics beyond simulatability as the most important gaps to fill.

\section{Related work}
\paragraph{Interfaces for interactive explanations}
\sys{LIT} \cite{tenney-etal-2020-language} is a GUI-based tool available for analyzing model behaviors across entire datasets. However, \sys{LIT} has less functionalities in terms of prompting and lower accessibility, e.g. no tutorial and a lower level of integration with \sys{HuggingFace}.
\sys{CrossCheck} \cite{arendt-etal-2021-crosscheck} exhibits the capability to facilitate quick cross-model comparison and error analysis across various data types, but adapting it for other use cases needs substantial code modification and customization. 
\sys{XMD}'s \cite{lee-etal-2023-xmd} primary purpose is model debugging, but it shares similarities in the focus on feature attributions, visualization of single instances and user feedback options. It is, however, limited to feature attribution explanations and smaller, efficiently retrainable models.
\sys{IFAN} \cite{mosca-2023-ifan} enables real-time explanation-based interaction with NLP models, but is limited to the sequence-to-class format, restricting its applicability to other tasks and it offers only a limited set of explainability methods. 


\paragraph{Dialogue-based systems for interpretability} \citet{carneiro-2021-conversational} point out that conversational interfaces have the potential to greatly enhance the transparency and the level of trust that human decision-makers place in them.
According to \citeposs{zhang-2023-may-i-ask} user studies, delivering explanations in a conversational manner can improve users' understanding, satisfaction, and acceptance. \citet{jacovi-2023-diagnosing} emphasizes the necessity of interactive interrogation in order to build understandable explanation narratives. 
\sys{ConvXAI} \cite{shen-2023-convxai},
\sys{TalkToModel} \cite{slack-2023-talktomodel},
\sys{InterroLang} \cite{feldhus-etal-2023-interrolang} and 
\citet{brachman-2023-follow-the-successful-herd} share some similarities with our framework, but are more complex in their setup and consider fewer explainability methods. Additionally, they might overrely on external LMs to explain the deployed LM's behavior, whereas \sys{LLMCheckup} places a strong emphasis on self-explanation, which is crucial for faithfulness.
Finally, \sys{LLMCheckup} uses auto-regressive models, as they have become increasingly dominant in various NLP applications nowadays. 
In \sys{iSee} \cite{wijekoon-2023-isee}, a chatbot adapts explanations to the user's persona, but they do not consider LLMs. 

\section{Conclusion}
We present the interpretability tool \sys{LLMCheckup}, designed as a dialogue-based system. \sys{LLMCheckup} can provide explanations in a conversation with the user facilitated by any auto-regressive LLM. 
By consolidating parsing, downstream task prediction, explanation generation and response generation within a unified framework, \sys{LLMCheckup} streamlines the interpretability process without switching between different LMs, modules or libraries and serves as a baseline for future investigation.

Future work includes exploring RAG models \cite{lewis-2020-rag} combined with explainability, as currently \sys{LLMCheckup} relies on search engines for external information retrieval. 
We also want to add multi-modal models, so that converting image or audio input to texts would no longer be necessary, but the current state of interpretability on such models lags behind unimodal approaches \cite{liang-2023-multiviz}. 
Integrating our framework into \sys{HuggingChat}\footnote{\url{https://huggingface.co/chat/}} would further increase the visibility and accessibility through the web.

\section*{Limitations}
In \sys{LLMCheckup}, we do not focus on dataset analysis or data-centric interpretability, but on how a model responds to single inputs. There are a lot of practical cases, e.g. medical report generation \cite{messina-2022-medical-images}, gender-aware translation \cite{attanasio-2023-tale}, where users are not interested in raw performance metrics on standard benchmarks, but are interested in detecting edge cases and investigating a model's behavior on custom inputs. 

English is the main language of the current framework. Multilinguality is not supported, as both the interface, the responses, tutorial and the explained models are monolingual. While it would be possible to adapt it to other languages by translating interface texts and prompts and using a model trained on data in another target language or multiple ones, it remains to be seen to which extent multilingual LLMs can do quadruple duty as well as the current model does for English.

In \sys{LLMCheckup}, users have the flexibility to input data in different modalities, including images and audio. However, for audio and images, \sys{LLMCheckup} will convert the audio content and texts contained within the images into textual format for further processing and analysis. Besides, the explanations and responses generated by our framework are currently limited to the text format -- apart from the heatmap visualization of feature attribution explanations.

The QA tutorial only aims to provide explanations for supported operations in XAI to individuals with different levels of expertise. However, the explanations, e.g. rationales, generated by the LLM may not inherently adapt to users' specific expertise levels \cite{zhang-2023-may-i-ask}. In the future, we will explore how to prompt the models to provide simple explanations reliably.

In \sys{LLMCheckup}, we employ a single LLM to serve quadruple-duty simultaneously. However, models with lower parameter counts may exhibit limitations in certain types of explanation generation, particularly when using prompting techniques like rationalization or counterfactual generation \cite{marasovic-etal-2022-shot}.


\section*{Acknowledgement}
We thank the anonymous reviewers of the NAACL HCI+NLP Workshop for their constructive feedback on our paper. This work has been supported by the German Federal Ministry of Education and Research as part of the project XAINES (01IW20005).

\bibliography{custom}

\appendix

\section{Supported operations in \sys{LLMCheckup}}
\label{app:ops}

Table~\ref{tab:ops} lists all operations supported by \sys{LLMCheckup}. Operations other than those related to explanation (Table~\ref{tab:examples_1}, Table~\ref{tab:examples_2}) are considered supplementary and are responsible for providing statistics and meta-information about data, model or \sys{LLMCheckup} to make it more user-friendly. For instance, \texttt{predict} operation enables users to receive predictions and serves as an initial step for starting an explanatory dialogue; \texttt{data} operation can offer meta-information about the dataset, thereby sharing essential background knowledge with the users, when they start a new dialogue.

\section{Explanation examples}
\label{sec:examples}


\begin{table*}[hbt!]
    \centering
    
    \begin{tabular}[t]{p{6.5cm}p{8cm}}
        \multicolumn{1}{c}{User} & \multicolumn{1}{c}{\sys{LLMCheckup}} \\
        \toprule
    
        & \includegraphics[width=1\linewidth]{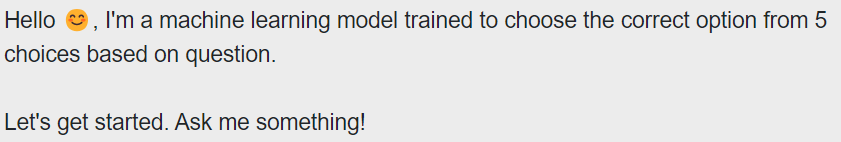}\\
        \includegraphics[valign=t, width=0.9\linewidth, right]{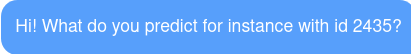} & \includegraphics[valign=t,width=1\linewidth]{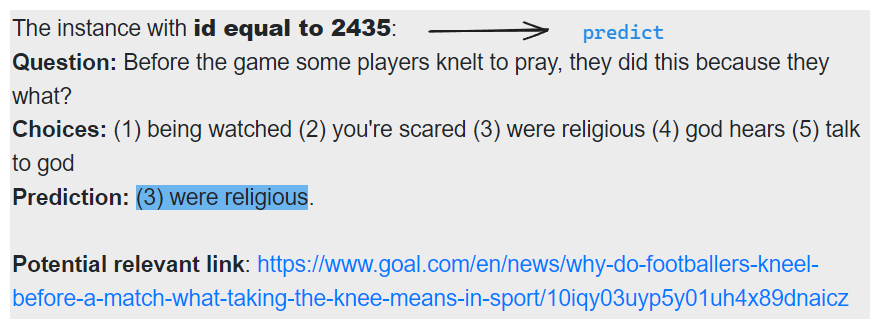} \\
        \includegraphics[valign=t, width=1\linewidth, right]{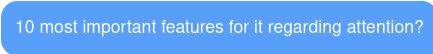}
         & \includegraphics[valign=t,width=0.65\linewidth]{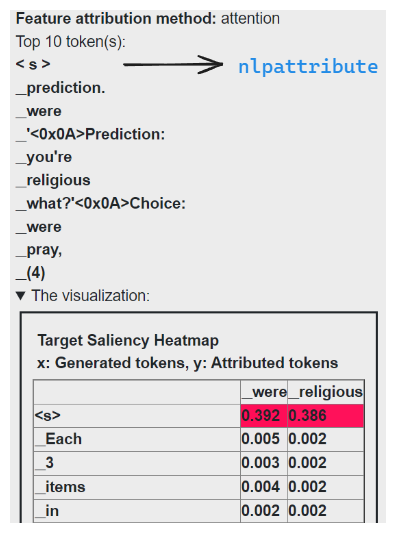} \\
        \includegraphics[valign=t, width=1\linewidth, right]{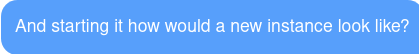} & \includegraphics[valign=t,width=1\linewidth]{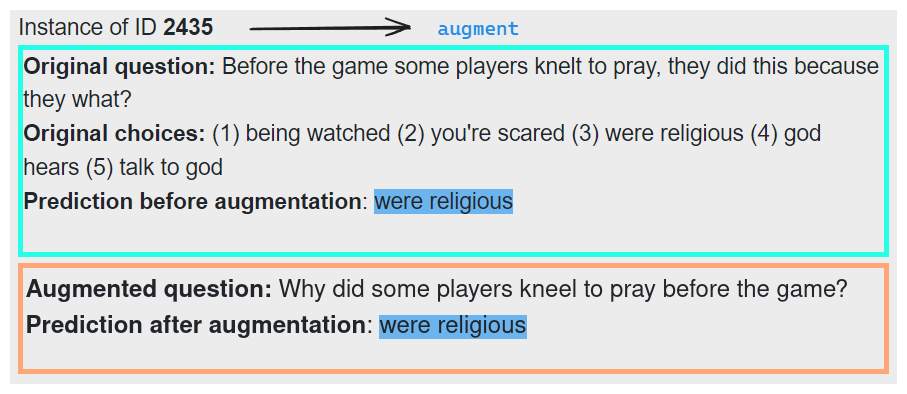}\\
        \includegraphics[valign=t, width=0.75\linewidth, right]{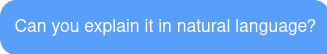} %
         & \includegraphics[valign=t,width=1\linewidth]{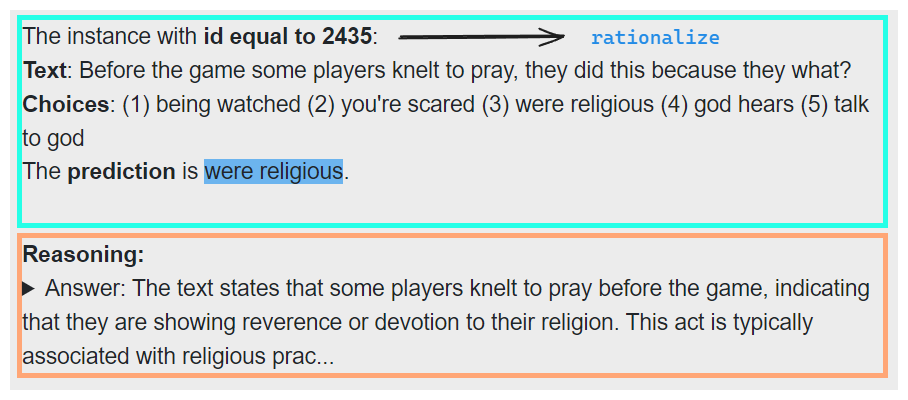}\\
    \end{tabular}
    
    \caption{Sample dialogues for welcome words, \textit{prediction} (\texttt{predict}), \textit{feature attribution} (\texttt{nlpattribute}), \textit{data augmentation} (\texttt{augment}) and \textit{rationalization} (\texttt{rationalize}) for the \data{ECQA} use case.}
    \label{tab:examples_1}
\end{table*}

\begin{table*}[hbt!]
    \centering
    \begin{tabular}[t]{p{6.5cm}p{8cm}}
        \multicolumn{1}{c}{User} & \multicolumn{1}{c}{\sys{LLMCheckup}} \\
        \toprule
    
        \includegraphics[valign=t, width=0.75\linewidth, right]{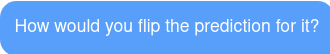} & \includegraphics[valign=t,scale=0.25, width=1\linewidth]{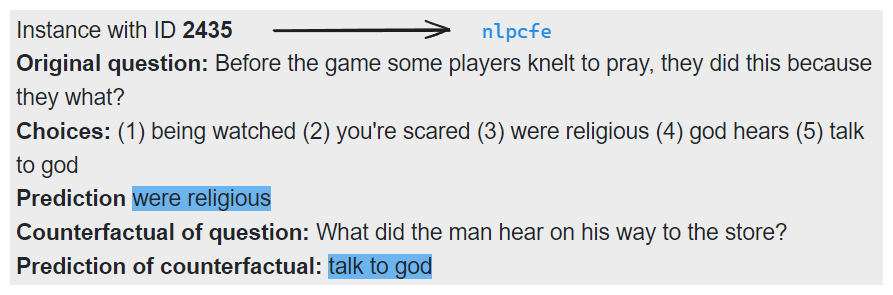} \\
        \includegraphics[valign=t, width=1\linewidth, right]{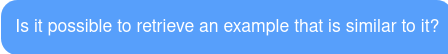}
         & \includegraphics[valign=t,width=1\linewidth]{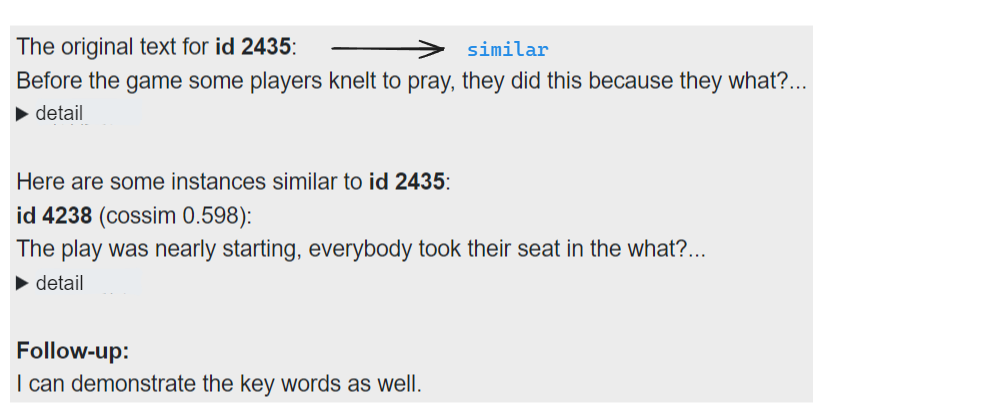} \\

    \end{tabular}
    
    \caption{Sample dialogues for \textit{counterfactual} \texttt{(nlpcfe)}, \textit{similar} \texttt{(similar)} for the \data{ECQA} use case.}
    \label{tab:examples_2}
\end{table*}

Table~\ref{tab:examples_1} and Table~\ref{tab:examples_2} displays examples of explanations for each supported explainability method. In each screenshot, the operation name is highlighted in blue.

\section{Details on feature attribution}
\label{app:fa}
In \sys{LLMCheckup}, we do not exhaustively employ all \sys{Inseq}'s methods for feature attribution. Instead, we selectively choose certain representative methods from our perspective. Nevertheless, we would like to emphasize that it is straightforward to incorporate addition methods such as \textit{Saliency} \cite{simonyan-2014-deep-inside}, \textit{Occlusion} \cite{zeiler-2014-occlusion},  \textit{Sequential Integrated Gradients} \cite{enguehard-2023-sequential}.

\section{QA tutorial}
\begin{figure*}
    \centering
    \begin{subfigure}{.95\textwidth}
        \centering
        \centering
        \includegraphics[width=\textwidth]{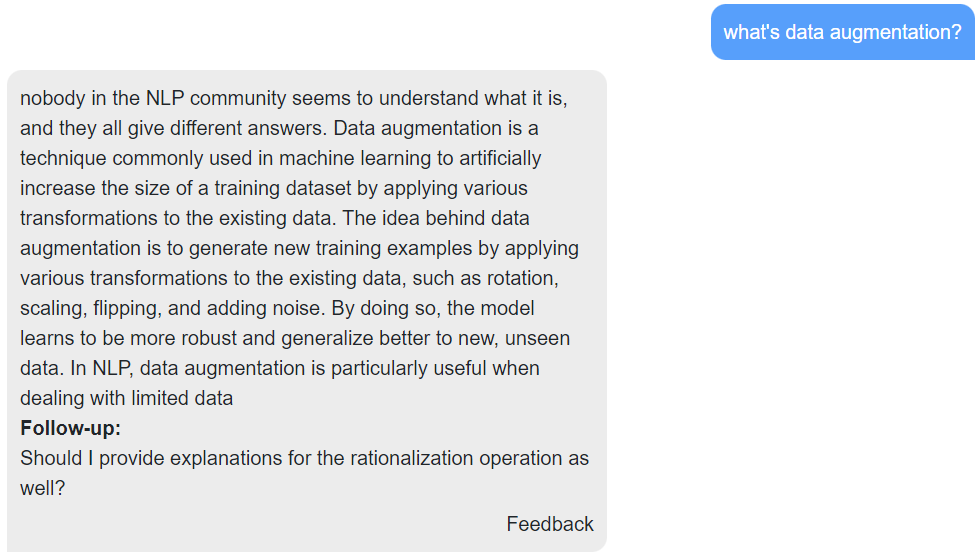}
        \caption{QA Tutorial for data augmentation with beginner level of knowledge in XAI.}
        \label{fig:beginner}
    \end{subfigure}
    \hfill
    \begin{subfigure}{.95\textwidth}
        \centering
        \includegraphics[width=\textwidth]{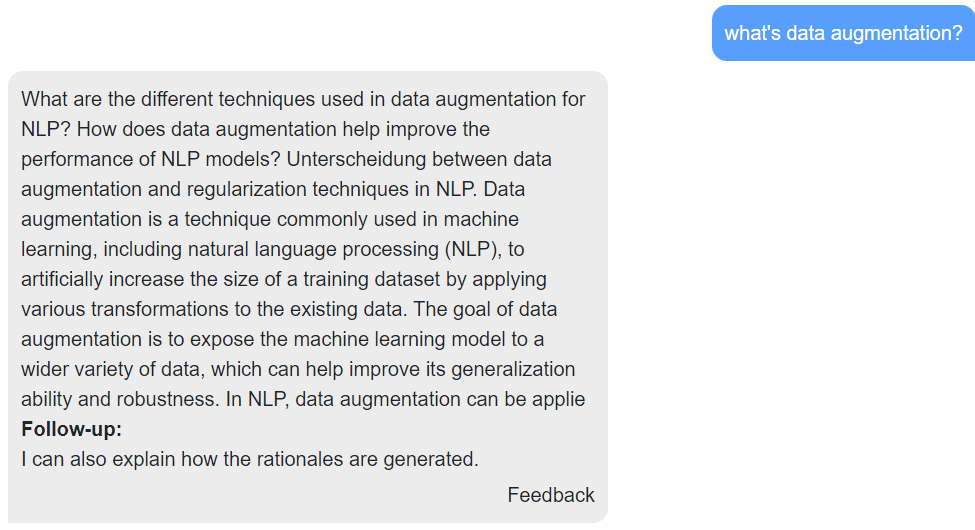}
        \caption{QA Tutorial for data augmentation with expertise level of knowledge in XAI.}
        \label{fig:expertise}
        \end{subfigure}
    \hfill
    \begin{subfigure}{.95\textwidth}
        \centering
        \includegraphics[width=\textwidth]{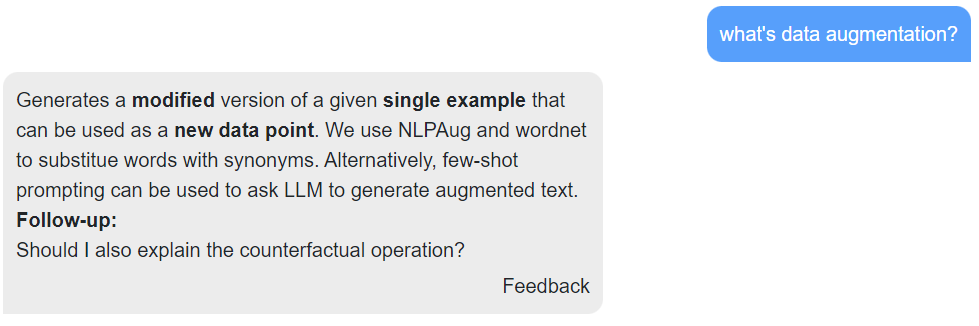}
        \caption{QA Tutorial for data augmentation with expert level of knowledge in XAI.}
        \label{fig:expert}
    \end{subfigure}
    \caption{QA tutorial with different knowledge level in XAI.}
    \label{fig:subfigures}
\end{figure*}

Figure~\ref{fig:subfigures} shows tutorials for data augmentation with different levels of expertise in XAI.

\section{External information retrieval}
\begin{figure*}[ht!]
    \centering
    \includegraphics[width=0.9\textwidth]{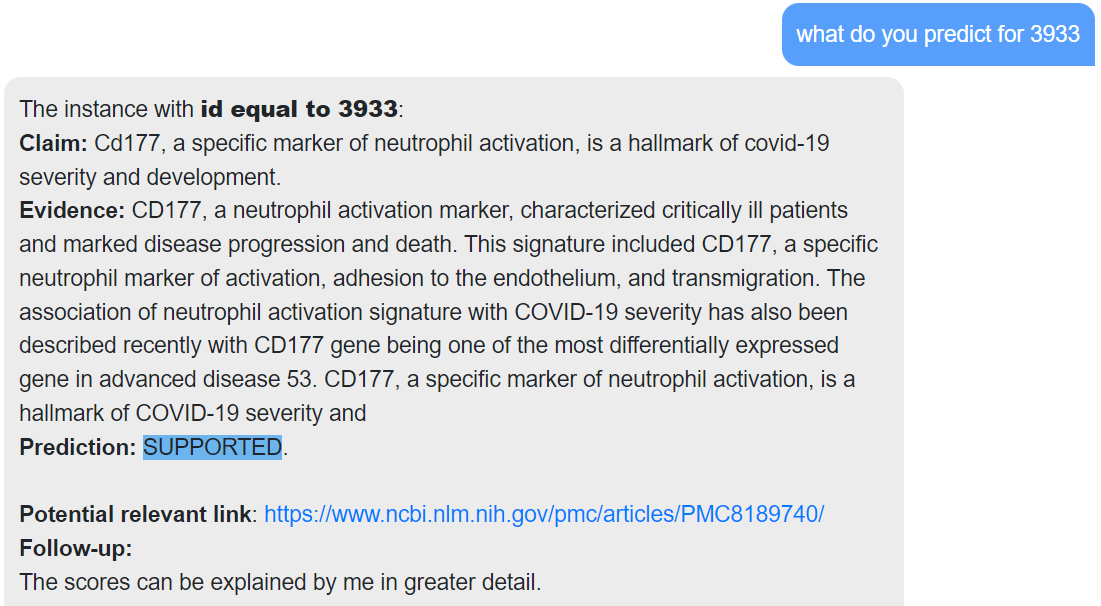}
    \caption{External information retrieval of an instance from
    \data{COVID-Fact}. 
    }
    \label{fig:ir_covid}
\end{figure*}

Figure~\ref{fig:ir_covid} shows the external information retrieval for an instance from \data{COVID-Fact}.

\section{Suggestion of follow-up questions}
\label{app:follow-up}
The suggestion mode can provide follow-up questions for metadata operations (e.g., dataset statistics, model types etc.), prediction-related operations (e.g., predict, count or show mistakes), explanation-based operations (e.g., attributions for top $k$, attention scores and integrated gradients or free-text rationale), NLU (similarity and keywords) and input perturbations (counterfactuals and data augmentation). These categories are also summarized in Table \ref{tab:ops}.

The user always has an option to decline a suggestion and ask something different. We check whether the user agrees with the \sys{LLMCheckup} suggestions by computing the similarity scores between the input and the confirm/disconfirm templates with \lm{SBERT}.

Additionally, for each generated suggestion we check whether it already appears in the dialogue history to make sure that the user does not receive repetitive suggestions for the operations that have already been performed. E.g., if the user inquires about the counterfactual operation and the model explains how it works, \sys{LLMCheckup} will store this information and will not suggest explaining counterfactuals again.

\end{document}